# Machine Learning Recommendation System for Health Insurance Decision Making In Nigeria

Ayomide Owoyemi[1,2], Emmanuel Nnaemeka[2], Temitope O. Benson[3,] Ronald Ikpe[4], Blessing Nwachukwu[1], Temitope Isedowo[2]

1. Department of Biomedical and Health Information Sciences, University of Illinois at Chicago.
2. Arteri Africa, Lagos, Nigeria
3. Institute for Computational and Data Sciences, University at Buffalo, State University of New York, Albany, USA.
4. Swansea University, United Kingdom

**Corresponding Author**

**Ayomide Owoyemi**

Department of Biomedical and Health Information Sciences

University of Illinois at Chicago. Chicago, USA

+13129782703

Aowoye3@uic.edu

**Abstract**

The uptake of health insurance has been poor in Nigeria, a significant step to improving this includes improved awareness, access to information and tools to support decision making. Artificial intelligence (AI) based recommender systems have gained popularity in helping individuals find movies, books, music, and different types of products on the internet including diverse applications in healthcare. The content-based methodology (item-based approach) was employed in the recommender system. We applied both the K-Nearest Neighbor (KNN) and Cosine similarity algorithm. We chose the Cosine similarity as our chosen algorithm after several evaluations based of their outcomes in comparison with domain knowledge. The recommender system takes into consideration the choices entered by the user, filters the health management organization (HMO) data by location and chosen prices. It then recommends the top 3 HMOs with closest similarity in services offered. A recommendation tool to help people find and select the best health insurance plan for them is useful in reducing the barrier of accessing health insurance. Users are empowered to easily find appropriate information on available plans, reduce cognitive overload in dealing with over 100 options available in the market and easily see what matches their financial capacity.

**RECOMMENDATION SYSTEM FOR HEALTH INSURANCE DECISION MAKING**

## 1. INTRODUCTION

Nigeria, a lower middle-income country, has a population of about 220 million people with a gross domestic product of 432 billion US dollars (1). Most of the healthcare funding coverage in Nigeria is out of pocket constituting over 70% of total healthcare expenditure in the country as at 2018, less than 5% of the population are covered under any form of health insurance exposing a large percentage of the population to financial risk and poverty (2). Ensuring financial protection and access to needed healthcare is integral to achieving Universal Health coverage (UHC) which is integral to the achievement of Sustainable Development Goal (SDG) 3.

The uptake of health insurance has been poor in Nigeria, and this has been due to a lot of challenges which include access to healthcare facilities, beliefs, low level of awareness about health insurance, policy challenges, poverty, and where to get required information (2–4). A significant step to improving this includes improved awareness, access to information and tools to support decision making (5). Recommender systems are designed to assist individuals to deal with a vast array of choices, it takes advantage of several sources of information to predict options and preferences around specific items (6–8). Recommender systems enhance the user experience by giving fast and coherent suggestions.

Artificial intelligence (AI) based recommender systems have gained popularity in helping individuals find movies, books, music and different types of products on the internet including diverse applications in healthcare (9–12). It has also been used in the insurance industry to support decision making on insurance products (13). Recommender systems are in three main categories which include: collaborative filtering, content-based and hybrid filtering (9). Collaborative filtering method uses the data from other users rating of items to make recommendation for a user for those items. There are two main types of collaborative filtering algorithms which are memory-based (or heuristic-based) and model-based, this is commonly used on video platforms to recommend popular content (14,15).

Content based recommendation methods use the data available about a specific user preference, search, or choice for a specific item in recommending similar items to that specific user, this is commonly used recommending items like personal computers, and mobile phones (16). Hybrid methods combine the two previous methods to avoid the limitations associated with the two methods, and they have been shown to provide more accurate recommendations than pure single methods (9,17). Examples of application of this include the systems used for Facebook friends' recommendation, Netflix movies recommendation and amazon product recommendation.

Health management organizations (HMOs) serve as agents of the National Health Insurance Scheme (NHIS) to offer health insurance cover to private and public sectors (18). There are presently 58 health management organizations (HMO) in Nigeria, accredited by the governing body for health insurance in Nigeria, and these 58 HMOs have an aggregate of 155 available plans for the public to select from. These plans vary based on price, benefits, geographical coverage, and value-added options (19). There is a complex decision-making process around selecting the appropriate HMOs coverage and what plan would be most appropriate for each person.

In this paper, we detail how we worked on developing a recommender system to improve decision making for Nigerians around choosing the most appropriate and suitable health insurance based on their personal preferences. To the best of our knowledge, no such system presently exists anywhere in the Nigerian insurance sector or in any form in the healthcare industry. This is a novel application of a recommender system in this region.

## 2. METHODS

The content-based methodology (item-based approach) was employed in the recommender system, which is significantly more accurate and efficient to use because the item-based method can be done offline and is non-dynamic, whereas the user-based method changes. This method can be implemented using the K-Nearest Neighbor (KNN) or Cosine similarity algorithm. Although the Cosine similarity was our chosen algorithm after several checks and comparison with the KNN algorithm. Let's briefly discuss both algorithms and their use cases.

### Cosine similarity

The Cosine similarity makes use of the cosine of angles between two vectors to check for most similar items to make its recommendations. Each record we have on the aggregated HMO dataset would be compared with a user's preference that has been pre-transformed to vectors.

The cosine between vectors can be calculated as:

$$\cos(\theta) = \frac{\mathbf{A} \cdot \mathbf{B}}{\|\mathbf{A}\|\|\mathbf{B}\|} = \frac{\sum\limits_{i=1}^{n} A_i B_i}{\sqrt{\sum\limits_{i=1}^{n} A_i^2}\sqrt{\sum\limits_{i=1}^{n} B_i^2}}$$

Cosine similarity is a value that is bound by a constrained range of 0 and 1. The similarity measurement is a measure of the cosine of the angle between the two non-zero vectors A and B. Suppose the angle between the two vectors A and B was 90 degrees. In that case, the cosine similarity will have a value of 0; this means that the two vectors are orthogonal or perpendicular to each other.

As the cosine similarity measurement gets closer to 1, then the angle between the two vectors A and B is smaller. This algorithm can be useful in finding the similarities between documents, in cases of plagiarism and Pose matching.

**KNN Algorithm**

The KNN implements the content-based technique by applying a distance-based algorithm, in our case we used the Euclidean distance.

$$= \sqrt{(q_1 - p_1)^2 + (q_2 - p_2)^2 + \cdots + (q_n - p_n)^2}$$

$$= \sqrt{\sum_{i=1}^{n} (q_i - p_i)^2}.$$

The nearest neighbors calculate distance between two vector spaces. Usually, the values range from 0 to infinity. So typically, the lesser the magnitude of the distance between the two vectors the more likely they are similar and therefore recommended by the algorithm. And the higher the magnitude between the vectors, the less likely those vectors are similar. The algorithm is popularly known to be applied in movie recommendations.

## 2.1 Steps taken to build our recommendation system.

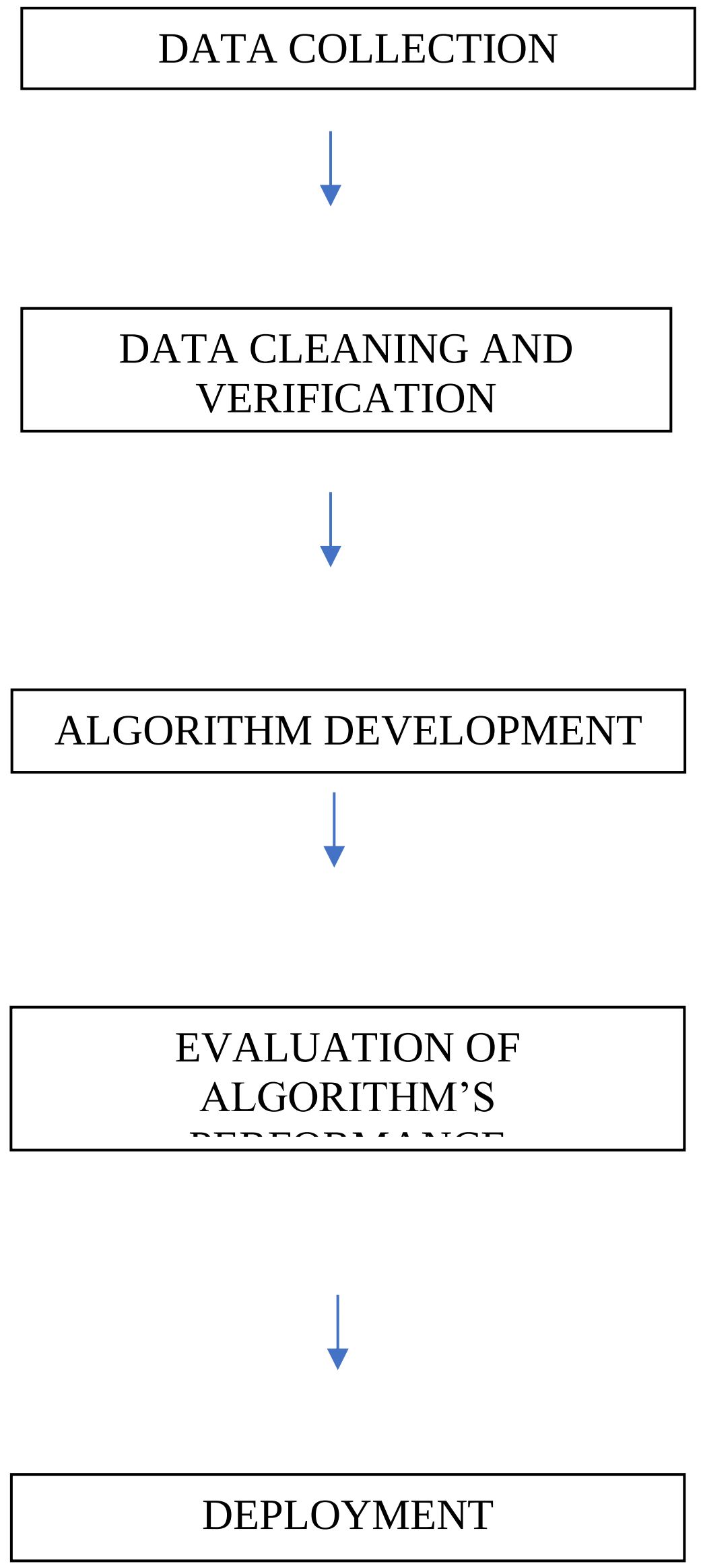

**Data Collection**

At the start of the project, the team came up with names of health insurance in Nigeria to make individual research on and collect those HMO's information on their individual plans and offer. About 148 health plans were highlighted and 14 features were extracted from the plans. A separate data on the Ratings on the HMOs from prospective clients was obtained via survey forms. This was used in the second phase of the recommender system. The included features are: Premium, geographical coverage, family Planning, Mental health, Dental care, Admission ward type, Telemedicine service, Cash back benefit, ANC delivery coverage, eye care cost limits, Gym membership, Annual Routine Medical Screening

**Data Cleaning**

The collected data was preprocessed and cleaned. In doing so, we considered the removal of certain attributes that seem to be redundant across all HMOs. We performed feature encoding on categorical features of the HMO data. Additionally, we performed feature engineering about the premium of all HMO, by creating four classes of premium based on its average pricing per plan since we were unable to get all their prices.

**Algorithm Development**

This took part in two steps:

1. Experimenting with KNN and Cosine similarity algorithms, with performance check done by our group of domain experts. After cleaning up our data and taking care of missing values and other related problems, we began experiments on two major algorithms, the KNN and the Cosine similarity. KNN applied the Euclidean distance in discovering the closest HMO recommended for a user based on their needs while Cosine similarity used the angles between two vectors (in this case, each HMO and our user's input) to decide which HMO would cater the most for our clients based on their inputted needs. For KNN, the smaller the distance between the user's choices and an HMO, the more suitable that HMO is to the user. This also applies to the cosine similarity, the smaller the angle between the two vectors (user choice and HMO services), the more suited the HMO is to cater for the user needs. It is also important to note that the algorithm was programmed to perform some low-level filtering for specific cases like location and amount preferred from clients' data input (both usually specified by the client upon selection of preference and needs) prior to any recommendation. This filter considers the users' location (currently Lagos State or nationwide for version 1 of the recommender system) and filters the HMO database such that no HMO outside the reach or location of the client would be recommended to them by the algorithm. It also applies to their amount remittable (ranging from tiers 1 to 4), and it works in a way, such that no tier higher than the user specified tier would be recommended. For instance, if a client chooses tier 3, the algorithm will only recommend HMOs that are within either tier 1, tier 2 or tier 3, excluding all tier 4 HMOs. This low-level filtering ensures all HMO services recommended to the users by the algorithm are all affordable (in terms of tiers) and accessible (in terms of location) to the user.

2. Using our Ratings for HMOs, we filtered through the suggested top recommendations from the algorithms and streamlined our recommendations to only three. After the experimentation in step 1 above, we proceed to filter through the recommendations from both algorithms individually and select the top 3 based on the ratings of the HMO obtained from our prospective clients. In this case, the algorithm might recommend 5 HMOs and their plans, we then select the best 3 using the data from the HMO ratings. This allows us to be able to select the people's choice alongside maintaining satisfaction of their needs. It is important to note that as we collect more candid data based on the ratings of the HMOs, the step 2 of the recommender system would function better, providing the people's choice to the people.

**Evaluation of algorithms performance**

Evaluating the performance of a recommender system can be quite tricky since it doesn't follow the conventional regression and classification ML problems. In evaluating our algorithm's performance, we employed the domain knowledge of our product using our experienced team of medical and domain experts. We curated several user's choices from the entire team of Arteri, we input these choices individually to the recommender system of both algorithms and came up with their individual recommendations. These recommendations were then vetted by the medical team to see which algorithm seemed to be suggesting the right HMO for the specific user needs. It was then collectively concluded that the Cosine Similarity gave the best recommendation as most users already on insurance plans got recommended to their HMOs after inputting their needs. This is to say that the cosine algorithm isn't so far away from the truth in recommending HMOs that are best suited for the user based on their needs.

**Deployment**

We then proceeded to the deployment stage of our recommender system. An Application Programming Interface (API) was built using the Fastapi framework on python. The API was then deployed to Heroku with the data running from GitHub and Postgres.

## 3. RESULTS

The user is expected to input their choice of HMO services, including their preferred location (Lagos state only and Nationwide) and payable amount (Tier1, Tier2, Tier3 and Tier4).

Figures 1 and 2, show the section where the user is required to enter their choice of preferred HMO services.

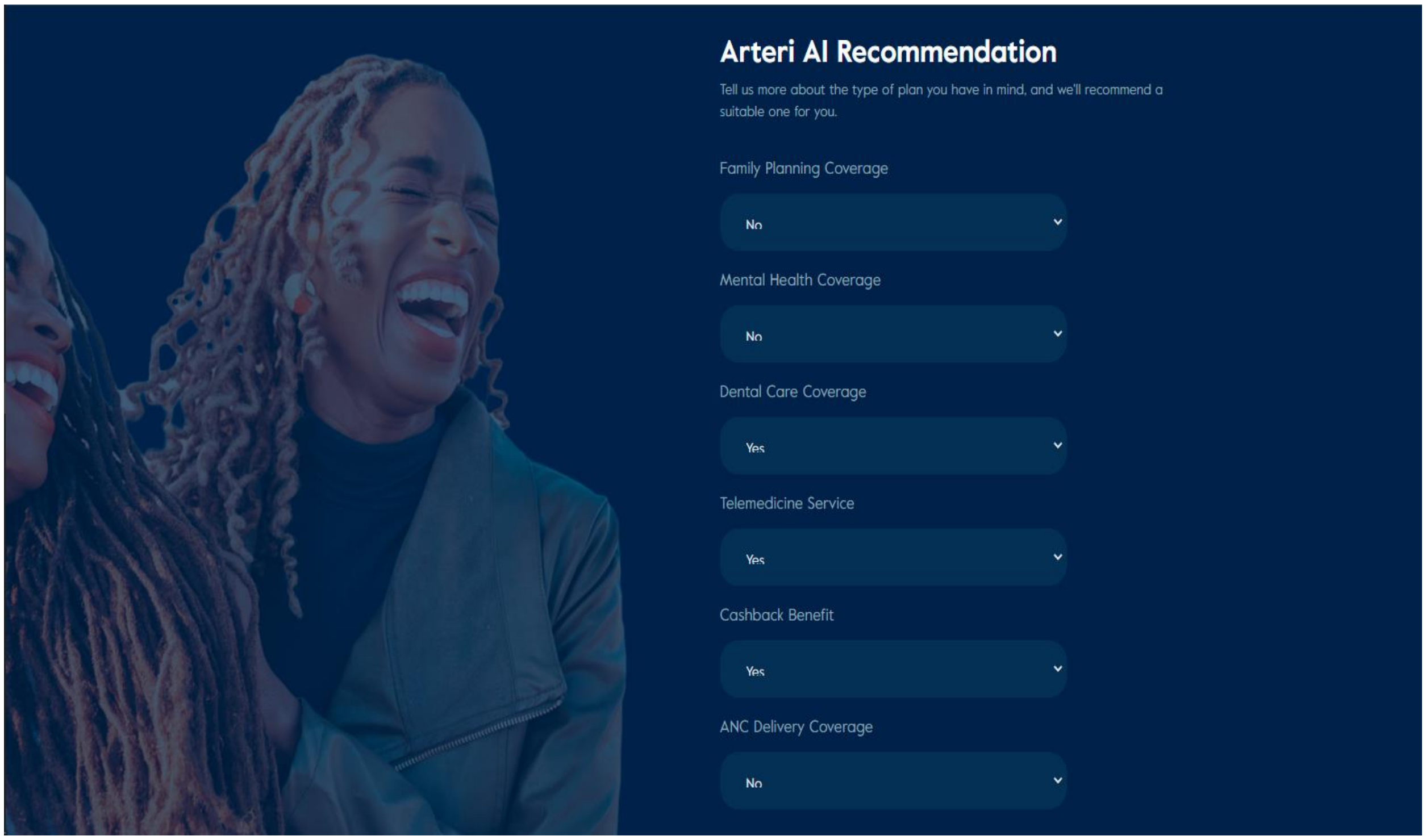

**Figure 1: The page where users enter their health plan preferences**

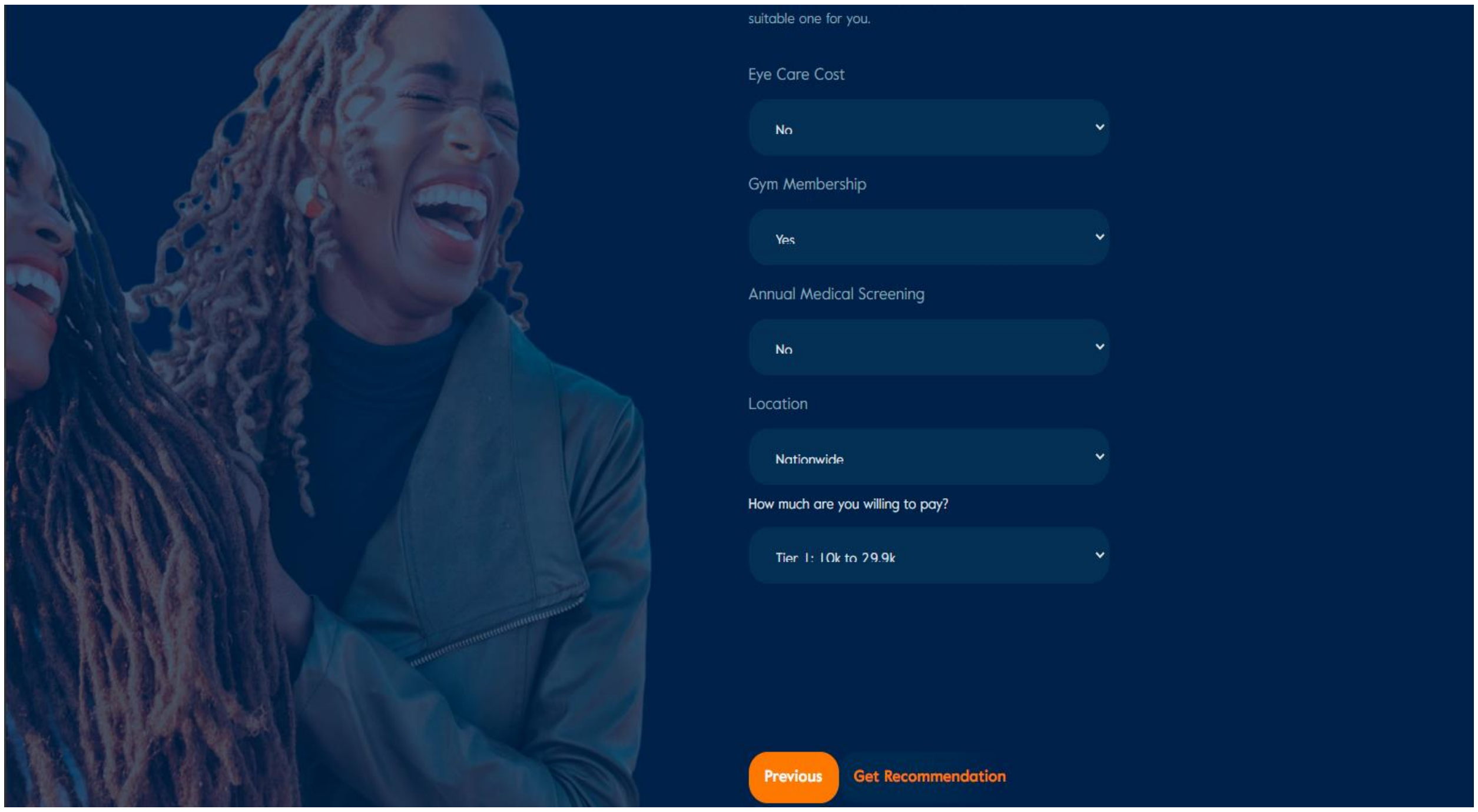

**Figure 2: Continuation of the page where users enter their health plan preferences**

The recommender system takes into consideration the choices entered by the user, filters the HMO data by location and chosen prices. It then recommends the top 3 HMOs with closest similarity in services offered as chosen by the users. The image (Fig. 3) below shows the top three HMO recommendations within Tier1(10k - 30k).

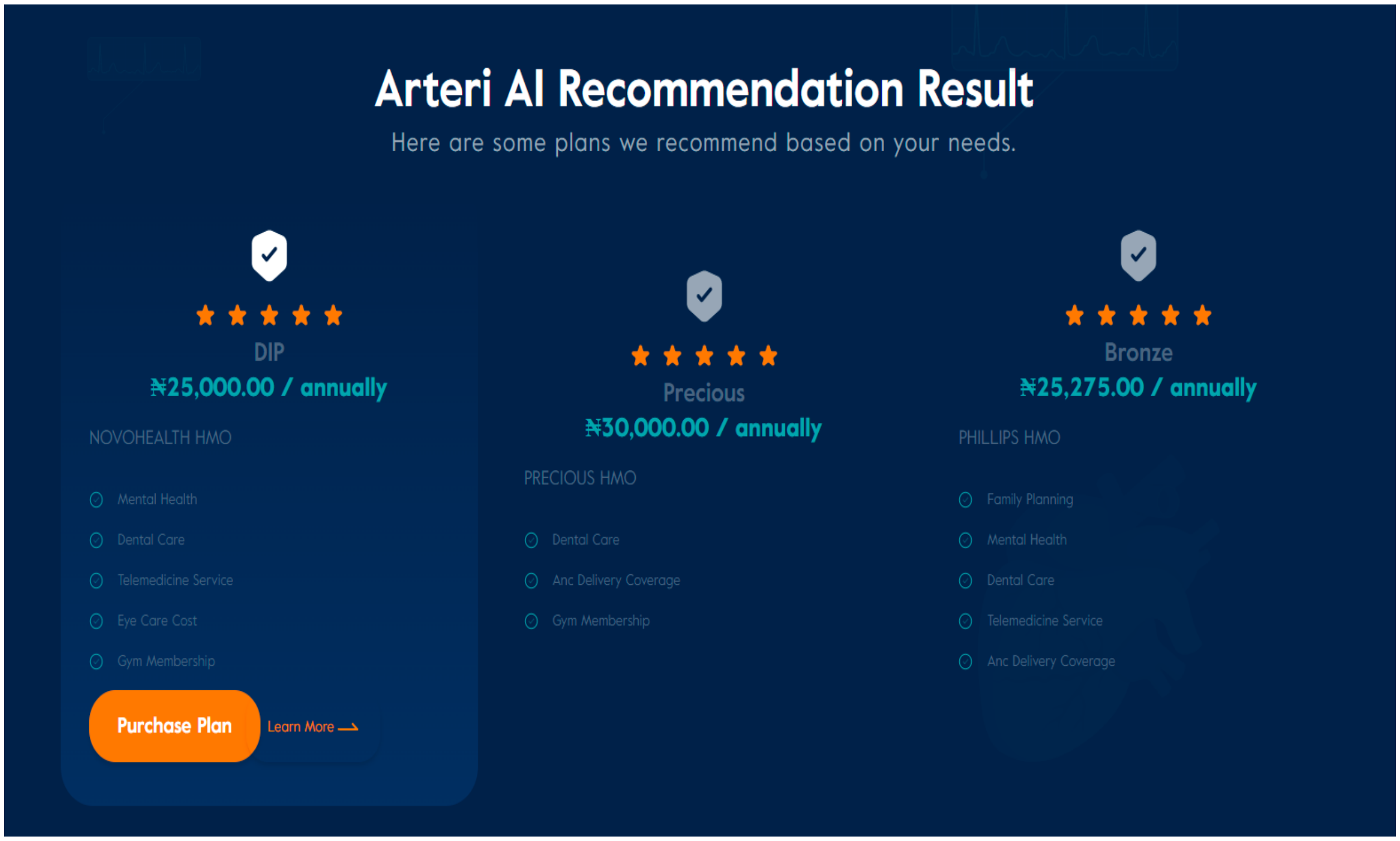

**Figure 3: The results page of the recommendation algorithm**

## 4. DISCUSSION

This is one of the first applications of recommender systems in healthcare and insurance that we know exist in Nigeria. The use of a tool to help people in efficiently making decisions about which health insurance product is most appropriate for them, will help in reducing the barrier of decision making for most Nigerians. Our recommender system offers to every user the top 3 health care insurance plans they can purchase based on their criteria and needs with an opportunity to complete the purchase and onboarding through the platform. For this purpose, we employed cosine similarity and KNN algorithm to recommend the results and propose three HMOs for users of the tool. Due to the difficulty in finding a ground truth algorithm and approach, we validated our results using medical professionals.

The tool was made available as a web page to ensure that anyone with access to the internet can easily access and use it without the challenge of having to download an app to their phones. As no other comprehensive database or information on available health insurance plans exists online for Nigerians, this platform helps people to find, choose and explore available health insurance plans. The tool also collects important data about users' health insurance choices and preferences which include, kinds of expected coverage in the health plan, special benefits like Gym membership, annual checkup, availability of telemedicine and what they are willing to pay for such options. This data is useful for planning for health insurance products, understanding preferences and demands and what most users are open to paying for specific coverages and benefits. We plan to incorporate more parameters and improve the recommender system by user behavior and other relevant data in such a way as to enhance our system and make it become more efficient to users both in Nigeria and the entire continent of Africa.

This study has various limitations which include the fact that there is mismatch between the granularity of the features users want and how the HMO display their services (which we use as features) on their site. Also, to make our data engineering easier we removed different features that were not straightforward and easy to encode, this would have led to the removal of some features that the users might need in making decisions. We also could not get the necessary data for some HMOs; therefore, we did not include them in the data used for the algorithm.

## 5. CONCLUSION

A recommendation tool to help people find and select the best health insurance plan for them is useful in reducing the barrier of accessing health insurance. Users are empowered to easily find appropriate information on available plans, reduce cognitive overload in dealing with over 100 options available in the market and easily see what matches their financial capacity.

Recommender systems are intended to assist users in making better decisions from a large content catalog. As a result, it focuses on the predictive algorithm's accuracy. The accuracy of a recommender system contributes to the user experience. Major benefits can be seen from utilization in how it enhances user experience, and easily analyze the market to discover user preferences and see what most people are interested in. This tool will indirectly contribute to improving the health insurance coverage in Nigeria and improve on the progress towards Universal Health Coverage for Nigerians.

## ABBREVIATIONS

| | |
|---|---|
| UHC | Universal Health coverage |
| SDG | Sustainable Development Goal |
| HMO | Health management organizations |
| NHIS | National Health Insurance Scheme |
| KNN | K-Nearest Neighbor |
| API | Application Programming Interface |

## DECLARATIONS

### Ethical Approval

No Ethical approval was required for this study.

### Consent for publication

All Authors give their consent for the publication of this article.

### Availability of data and material

N/A

### Competing Interests

AO, EN and TI are employees of Arteri Africa. The other authors have no competing interests to declare.

### Funding

The authors did not receive support from any organization for the submitted work.

### Authors Contributions

All authors contributed to the study conception and design. Material preparation, data collection and analysis were performed by Ayomide Owoyemi, Emmanuel Nnaemeka, Ron Ikpe & Temitope Isedowo. The first draft of the manuscript was written by Ayomide Owoyemi, and all authors commented and edited subsequent versions of the manuscript. All authors read and approved the final manuscript.

**Acknowledgements**

We would like to acknowledge the efforts and assistance of Seyi Morountonu and Olukoya Fatosa in the conception and the execution of this project.

**Ethical Statement**

The authors are accountable for all aspects of the work in ensuring that questions related to the accuracy or integrity of any part of the work are appropriately investigated and resolved.

**Figure Legends**

Figure 1: The page where users enter their health plan preferences

Figure 2: Continuation of the page where users enter their health plan preferences

Figure 3: The results page of the recommendation algorithm